\documentclass[a4paper, conference]{ieeeconf}

\IEEEoverridecommandlockouts          
\usepackage{graphicx}
\usepackage{amsmath}
\usepackage{mathabx}
\usepackage{algorithm}
\usepackage{algpseudocode}
\usepackage[T1]{fontenc}
\usepackage{array}
\usepackage{booktabs,ragged2e}
\usepackage{float}
\usepackage[flushleft]{threeparttable}
\usepackage{color,soul}

\algnewcommand\algorithmicforeach{\textbf{for each}}
\algdef{S}[FOR]{ForEach}[1]{\algorithmicforeach\ #1\ \algorithmicdo}

\title{\LARGE \bf
 A Data-Efficient Model-Based Learning Framework for the\\ Closed-Loop Control of Continuum Robots
}

\author{Xinran~Wang,~\IEEEmembership{Student~Member,~IEEE} and Nicolas~Rojas,~\IEEEmembership{Member,~IEEE}
\thanks{Xinran Wang and Nicolas Rojas are with the REDS Lab, Dyson School of Design Engineering, Imperial College London, 25 Exhibition Road, London, SW7 2DB, UK
{\tt\small (xinran.wang20, n.rojas)@imperial.ac.uk}}%
}

\begin{document}

\maketitle
\thispagestyle{empty}
\pagestyle{empty}
\graphicspath{ {images/} }
\begin{abstract} 

Traditional dynamic models of continuum robots are in general computationally expensive and not suitable for real-time control. Recent approaches using learning-based methods to approximate the dynamic model of continuum robots for control have been promising, although real data hungry---which may cause potential damage to robots and be time consuming---and getting poorer performance when trained with simulation data only. This paper presents a model-based learning framework for continuum robot closed-loop control that, by combining simulation and real data, shows to require only 100 real data to outperform a real-data-only controller trained using up to 10000 points. The introduced data-efficient framework with three control policies has utilized a Gaussian process regression (GPR) and a recurrent neural network (RNN). Control policy A uses a GPR model and a RNN trained in simulation to optimize control outputs for simulated targets; control policy B retrains the RNN in policy A with data generated from the GPR model to adapt to real robot physics; control policy C utilizes policy A and B to form a hybrid policy. Using a continuum robot with soft spines, we show that our approach provides an efficient framework to bridge the sim-to-real gap in model-based learning for continuum robots.
\end{abstract}



\section{Introduction}
\begin{figure*}[t!]
 \centering
 \includegraphics[width=\textwidth]{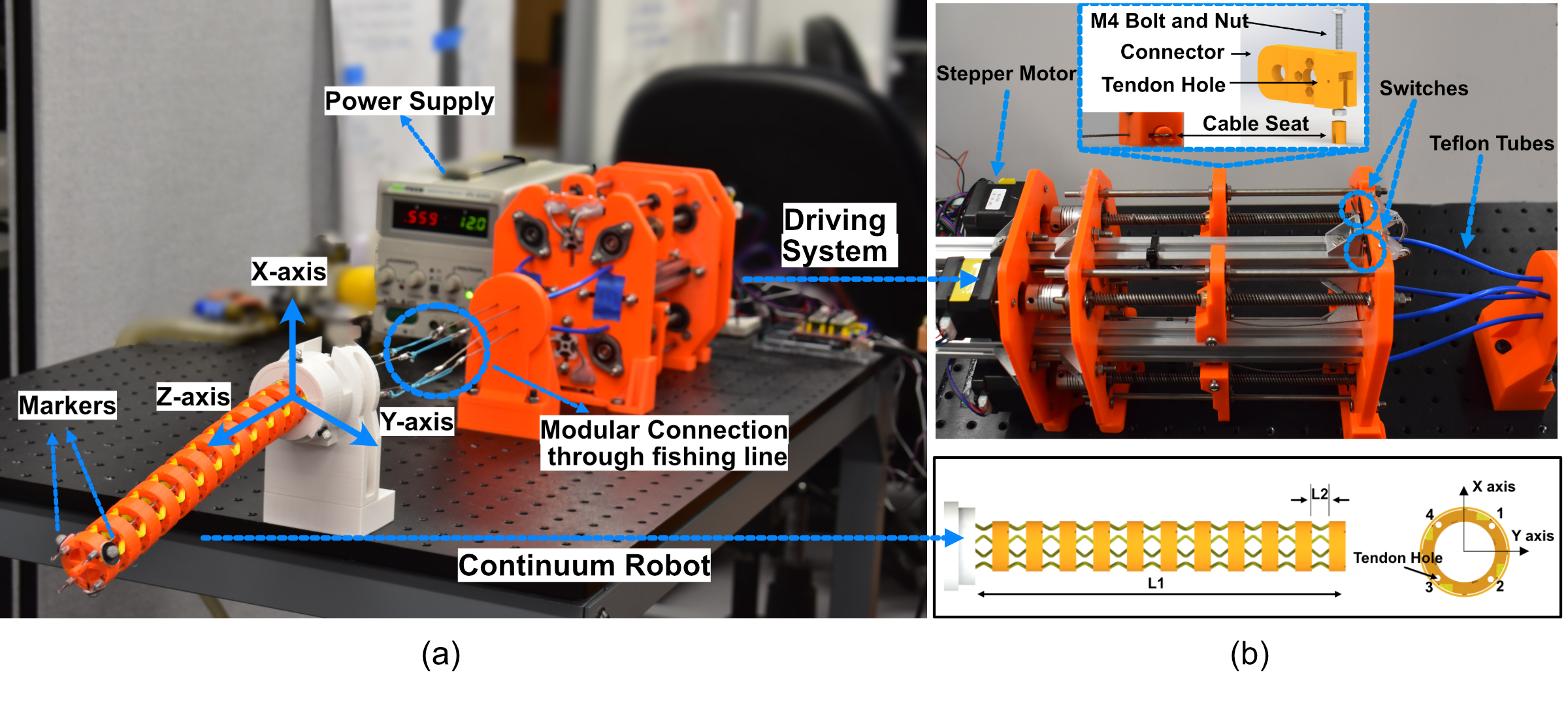} 
 \caption{(a) Overview of the experimental setup. (b) Details of the driving system and continuum robot design.}
 \label{fig:3}
\end{figure*}
For controlling continuum robots, state-of-the-art numerical models like constant curvature models, Beam theory, and Cosserat rod theory are commonly used \cite{renda2018discrete}\cite{gravagne2003large}\cite{cc2010}. Constant curvature (CC) models assume the deformed backbone can be represented as a planar constant curvature shape and ignore external load effects to achieve a low computation cost control; piece-wise constant curvature models are later developed for multiple section continuum robots \cite{cc2010}. The Cosserat rod dynamic model has higher accuracy and fidelity when considering external load over the CC approach \cite{benchmark_modelling}. However, while there exist attempts to perform real-time control using the Cosserat rod theory \cite{till2019real}, the Cosserat rod approach is computationally expensive. Indeed, due to the non-linear characteristic of soft materials, external uncertainties, and expensive computation costs of traditional numerical methods, learning-based approaches are deployed to achieve real-time control of continuum robots \cite{wang2021survey}.

Machine learning methods like Gaussian Process Regression (GPR) and Kalman filter have been used in controlling continuum robots. For example, Fang et al. present a model-free online learning method using multiple local Gaussian process models to represent an inverse mapping of soft continuum robots using real data~\cite{fang2019vision}. Previously, it has been shown that GPR can achieve data efficient control in different robot applications \cite{deisenroth2011pilco}; and more recently, a method using GPR to model time-series dynamics alongside deformation of soft robots with shape memory actuators was presented in \cite{sabelhaus2021gaussian}. The GPR-based methods in \cite{fang2019vision} and \cite{sabelhaus2021gaussian} are a move forward for model-free techniques in terms of data efficiency for continuum robot control, but both require several thousand of real data points to deliver results---thus limiting their real-time deployment in changing environments. Regarding Kalman-filter-based methods, these usually require high frequency readings to get real-time control, such as the approach to perform path tracking of a continuum robot presented in \cite{li2017model}---which also neglects robot dynamic effects.

Other model-free data-driven learning methods, like deep q-learning, utilize neural networks and gradient descent searching for getting an optimal policy to control continuum robots~\cite{model_free}\cite{satheeshbabu2019open}\cite{wu2020position}\cite{jiang2021hierarchical}. Some drawbacks of these methods include a simulation-to-real gap that leads to unsatisfactory performance in the real robot and the need for large numbers of real data points to get good results. Model-based learning methods have also been proposed, which mainly relying on training a model representation (e.g., \cite{moerland2020model}\cite{della2021model}). For instance, a feedforward neural network is commonly used in quasi-static control of soft robots~\cite{bern2020soft}, while a recurrent neural network is commonly seen in dynamic modelling of these manipulators~\cite{chikhaoui2018control}\cite{thuruthel2018model}. For modelling dynamics of a system using model-based learning method, it is found that training directly from real data is better than from simulation data, which requiring above 7000 real data points to achieve a desired performance\cite{thuruthel2018model}. 

A first-order dynamic control of soft continuum robots, which also uses supervised learning to train a dynamic model representation using real experiments data, is presented in \cite{george2020first}. A simple control policy is trained reusing obtained real experiment data to perform the learning by skipping trajectory optimization steps. However, the resulting supervised learning method is not data efficient, requiring also >7000 data points. Long-time training using real experiment data is troublesome as it leads to potential damage to robots, is time-consuming, and is not robust against external environment changes. In contrast, a data-efficient approach benefits from rapidly performing calibration after external environment changes.

Beyond using supervised learning to improve the accuracy of the trained model, adaptive control and iterative control have been also used as complementary methods to model-based learning when acquiring new information from feedback control or feedback action \cite{wang2021survey}\cite{della2021model}\cite{cao2021observer}\cite{tang2019novel}. However, these methods rely on continuous real data feedback to gradually improve the controller's performance or recording multiple data at the same points, which limits heavily their data efficiency.

This paper presents a data-efficient framework to model the dynamics of tendon-driven continuum robots, which may be affected by deformation of a compressible spine. A recurrent neural network is first trained on simulation data obtained from a Cosserat rod dynamic model, adapting it to use tendon length control instead of tension force control \cite{till2019real}\cite{continuum_tendon_length}. We also develop a compressible spine model for accurately modelling tendon length changes in these cases. The error and uncertainties between the real continuum robot and the simulated dynamic model are modelled via a Gaussian Process Regression using sparsely picked pre-defined points in the real workspace. 

 Both open-loop and closed-loop control policies were tested on a physical tendon-driven continuum robot with compressible spine for control policy A and B. And control policy C is tested in closed-loop control. The results show that our approaches achieves comparable or better performance, using only 100 real data points, than controllers trained with the same RNN architecture directly from up to 10000 real data points. To the best of our knowledge, our approach has achieved the best data-efficient modelling for continuum robot control.

The rest of this paper is organised as follows. Section II introduces the continuum robot with compressible spine, along with its driving system, that is used in our experiments. Section III discusses  the compression model developed for the compressible spine and the calibration procedures of the theoretical dynamic model. Section IV describes our model-based learning methods for the continuum robot and proposed control policy A, B and C. Section V presents our experimental results and discussion. Finally, section VI concludes the paper.

\section{Continuum Robot with Compressible Spine}
The continuum robot used for the experiments herein reported (Fig.~\ref{fig:3}(a)) has four compressible spines, based on the spine design presented in \cite{clark2020design}. We 3D-printed the spine of the continuum robot using thermoplastic polyurethane and embedded them inside 12 disks, leaving us 11 compressible sections. There are four tendon paths as shown in Fig.~\ref{fig:3}(b). Four spines are embedded in each disk. The overall length of the continuum robot $L1$ is 0.22 m, and each compressible section of it has a length $L2$ of 0.01 m in a relaxed state. This type of continuum robot is not expensive, and is easy to manufacture and replace. It is inherently safe due to its flexibility and compliance of the compressible spines and low stiffness compared to traditional continuum robots while able to pass large cables or objects through its center.

The design of the driving system comes with four stepper motors activating four lead screws. Four switches act as resetting points in front of a 3D-printed plate. A specific connector was designed to connect tendons and lead-screw as shown in \ref{fig:3}(b). The connector was 3D printed, also acting as a cable tensioner. M4 bolts and nuts in the connector are used to push the cable seat, reducing the overall cable length and increasing tendon tension while holding it to places. The tendons are steel-wire with in-compressible Teflon tubes used as guiding paths to continuum robot.

The overall continuum robot with the tendon-length control driving system is fixed on a metal breadboard. On the tip and base of the continuum robot, there are a total of six markers for motion tracking. In addition, six OptiTrack motion tracking cameras are set up in the room with a tracking frequency of 100Hz. The averaged tracking error in each axis is less than 0.5 mm. The continuum robot has modular connections using fishing line to remove from or attach to the driving system easily.

\section{Mathematical Model}
For modelling our continuum robot, we have adapted the tendon-driven dynamic Cosserat rod model in \cite{till2019real}\cite{rucker2011statics}\cite{janabi2021cosserat}. More discussion about using tendon-length control are presented in \cite{till2019real}\cite{continuum_tendon_length}. In this section, we mainly talked about the modification added to tendon-driven dynamic Cosserat rod model regrading the compressible spine used in \cite{till2019real}.

\subsection{Compression Model for the Soft Spine}
Along the centre-line of the rod, we have defined the position $p(s)$, orientation $R(s)$, the linear velocity $v(s)$, angular velocity $u(s)$, internal force $n(s)$ and internal moment $m(s)$ at the length s of the rod. And $q$ is linear velocity  and $\omega$ is angular velocity  for the rod's time $t$ at length $s$. And $p_s$ represents taking derivative of position respect to length of the rod s and so on. For the Cosserat rod model, most internal forces come from tendon tensions, which are compression forces. Under the condition of little external load and lightweight of the continuum robot, we have considered the internal forces compress the spine.

With this assumption, multiple trials of experiments have been conducted to determine the relationship between the length change of our compressible soft spine with compression force. A specimen of two sections of the compressible spine is used for testing. The change of total length of the specimen is recorded with respect to total forces applied on top of the specimen. Finally, we fit all the experiment data points using a regression method and simplify it into a linear equation, a decrease of compressible spine length with respect to force is represented as :
\begin{align}
    \begin{split}
        l_c&=c_{spine}\times{\left \|  n(s)\right \|},\\
     \end{split}
     \label{eq:14}
\end{align}
where $c_{spine}$ is the scalar coefficient obtained from experiments with a value of $0.2\frac{mm}{N}$ for each section of the compressible spine, and $\left \|  n(s)\right \|$ is the magnitude of vector $n(s)$ at length $s$ of the rod. The length stops decreasing after compression force exceeding 20 Newtons.

Our continuum robot has 11 sections of compressible segments as shown in Figure \ref{fig:3}(b). Meanwhile, internal forces $n(s)$ are different at these 11 sections due to weight and external load effects are calculated using linear constitutive law:
\begin{align}
    \begin{split}
        n(s) & =R[K_{se}(v-v^*)],   
    \end{split}
    \label{eq:30}
\end{align}

where $K_{se}$ is stiffness matrix for shear and extension and $v^*$ is the straight rod reference configuration with a value of [0;0;1].
Therefore, instead of directly reducing the total length of the continuum robots, we have iteratively reduced the length on each segment of the continuum robots with respect to internal force at the length s of the rod inside the shooting method of the Cosserat rod model. The detailed algorithm is presented in Algorithm 1.
\begin{algorithm}[t]
\caption{Calculating overall reduced length of the continuum robot with compressible spine}
\label{alg:1}
\begin{algorithmic}
\While{Error >  0}
\State Solving system ODEs for $p_s, R_s, v_s, u_s, q_s, \omega_s$
\State Using Euler's method to calculate $p,R,v,u,q,\omega$
\ForEach{ compressible section i, i = 1,2,...,11}
\ForEach{ Spatial Point j inside section i}
\State Calculate $n(i)$ at segment $i$ with Equation \eqref{eq:30}
\State Calculate $l_c(i)$ using Equation \eqref{eq:14}
\State Update $ds(i)$ as $ds(i)=ds(i)-l_c(i)$
\State Update $p(j),R(j),v(j),u(j),q(j),\omega(j)$ using Euler's Integration method
\EndFor
\EndFor
\EndWhile
\end{algorithmic}
\end{algorithm}
\subsection{Calibration of Mathematical Model}
We have used four 3D-printed Thermoplastic Polyurethane (TPU) spines in the continuum robot. Meanwhile, the Cosserat mathematical model is based on a single rod. We treat these four separate spines as one centre TPU spine with the same cross-section area to simplify our model. Meanwhile, the spine has triangular shapes along its length, resulting changes in its mechanical propriety. Therefore, we need to re-calibrate Young's modulus ($E$). There is an existing relationship between Young's modulus and shear modulus ($G$) for isotropic and homogeneous materials, which is $G=E/(2\times(1+v)$, where $v$ is Poisson's ratio. For the convenience of calibration, we apply this existing relationship and calibrate Young's modulus of the spine. 

The Poisson's ratio value is set to a fixed value of 0.3897 obtained from \cite{lee2019evaluation}. We have performed experiments by pulling tendons using different weights and recording tip position values using motion-tracking cameras. To calibrate, we solve an optimisation problem of minimising the error terms between experiment data and simulated mathematical data by finding suitable Young's Modulus of the compressible spine. The final calibrated Young's Modulus value is $2.33e^9$. 

\section{Model-Based Learning framework and Control}

\begin{figure*}[t!]
 \centering
 \includegraphics[width=\textwidth]{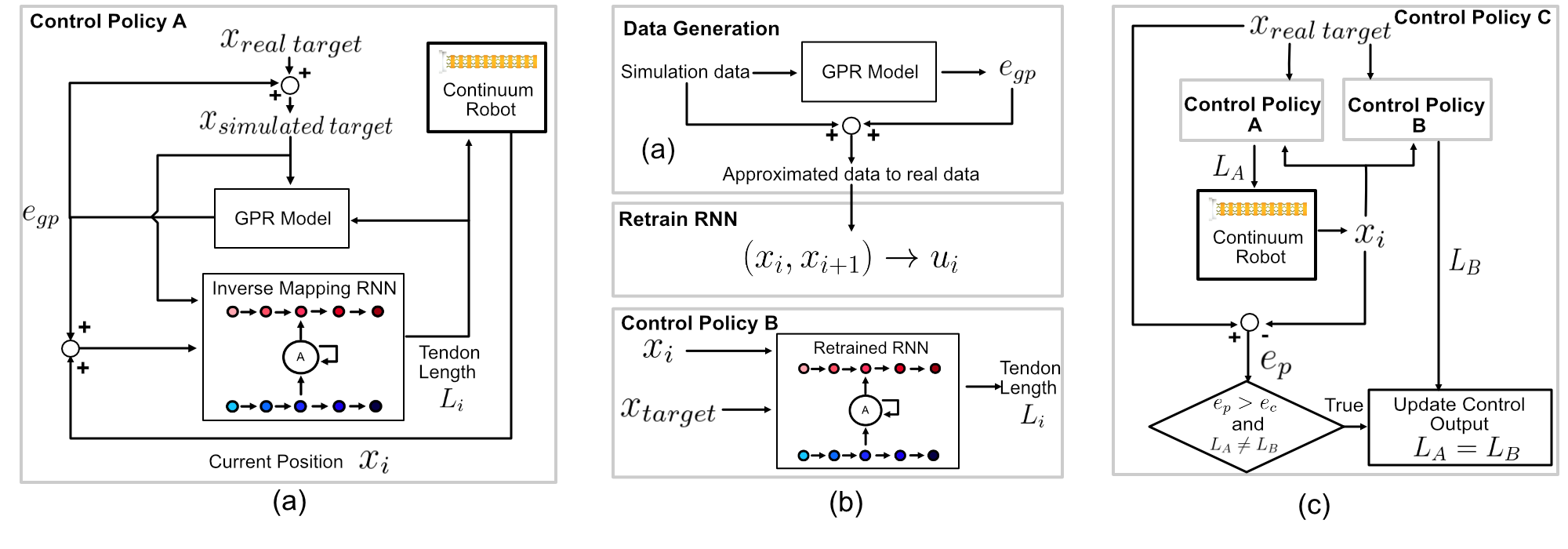} 
 \caption{(a) control policy A: Control diagram of using GRP model and RNN trained in simulation (b) control policy B: Retraining RNN in control policy A with GPR model generated data to obtain a controller. (c) control policy C: A hybrid control policy using both control policy A and B. This policy mainly relies on control policy A, with the correction from control policy B when position error is large.}
 \label{fig:7}
\end{figure*}

The model-based learning framework has utilized two main components to form hybrid controllers and achieve data efficiency: one is a RNN to model inverse dynamics of the continuum robot in simulation. The other component is a Gaussian process regression model to compensate for error between simulated and real continuum robots. In this section, we present the detailed modelling methods and three control policies based on this framework. The detailed block diagrams are shown in Fig. \ref{fig:7}. 
\subsection{Recurrent Neural Network Model}

The dynamic modelling of soft continuum robots was presented previously in \cite{thuruthel2018model}. In their approach, a fully dynamic model with $n$ degree-of-freedom in configuration-space variable q is represented as:
\begin{align}
    \begin{split}
        M(q)\ddot q+C(q,\dot q)\dot q +G(q) =\tau\\
     \end{split}
     \label{eq:15}
\end{align}
where $M(q)$ is the inertial term, $C(q,\dot{q})$ is the damping term and $G(q)$ is the gravitational and stiffness effects. $\tau$ is generalized force acting on the system. This representation is later reduced to a first-order dynamic neural network in \cite{george2020first}. In the first-order dynamic representation, they considered the inertial term mainly takes effects during initial transient motion, and most soft continuum robots are lightweight. The first-order dynamic model is then represented as:
\begin{align}
    \begin{split}
        C(q,\dot q)\dot q +G(q) & =\tau\\
        C(q,\dot q)\dot q +G(q) &=\tau - B^T(q)u\\
    \end{split}
    \label{eq:16}
\end{align}

For the first-order recurrent neural network, $\dot{q}$ is provided and feed into the input. Regarding under-actuated systems actuated by $m$ control inputs $u$, a $B^T$ describing the  $n\times{m}$ matrix of control input $u$ influences on $\tau$  is included in Equation \eqref{eq:16} \cite{underactuated_sys}. However, during experiments, changing positions are more easily recorded. In their approach, fixed time-step positions are recorded and fed into the neural network to represent discretized dynamics. The mapping in the RNN is now represented as:
\begin{align}
    \begin{split}
        (u_i, x_i)\rightarrow x_{i+1}\\
    \end{split}
    \label{eq:17}
\end{align}

where $u_i$ is the control input in the under-actuated system and $x_i$ is the position at time-step $i$.

We followed the approach of first-order dynamic recurrent neural network modelling. From the tendon-length control Cosserat dynamic model, we generated 10000 data points using the continuous random exploration of the actuators with a maximum travel distance of 0.003 m each step. The tendon length control is slower than pneumatic pressure control. By imposing the maximum travel distance so it can capture the dynamic effects under the assumption of 10 Hz control frequency. In addition, to not restricting the mathematical model from exploring the workspace, we have two tendons actuating each time. We are only minimizing the tendon length errors of two actuating tendons leaving the remaining two tendon lengths calculated using Euler's integration method along their tendon paths.

The neural network consists of a single-layer recurrent neural network of 30 neurons with a linear output layer. We divided 10000 data points with a ratio of 70:30 for the training and validation dataset. To prevent our model from over-fitting, we have adapted the early stopping method. With the training of 780 epochs, the validation mean absolute error is around 0.0024 meters.

\subsection{Gaussian Process Regression}
For modelling the error between our recurrent neural network and the real continuum robot with data efficiency, we employed a non-parametric method GPR. Now we consider an input data $X={x_i}$ and output data $Y={y_i}$ for i = 1,2,...,N. GPR could approximate any nonlinear function $y_i\sim f(x_i) + \epsilon$, where $\epsilon$ is a white Gaussian noise with zero mean and variance $\sigma^2_n$ \cite{rasmussen2006cki}. And $f(x_i) \sim GP(m(x_i), k(x_i,x'))$. The mean function and $k(x_i,x')$ defined the GPR. For our application, we have chosen the zero-mean function and square exponential (SE) kernel. The SE kernel is represented as:
\begin{align}
    \begin{split}
        k(x_i,x') = \sigma^2_sexp(-\frac{1}{2}(x_i-x')^T\lambda(x_i-x'))
    \end{split}
    \label{eq:18}
\end{align}
where $\sigma^2_s$ is the signal variance, $\lambda:= diag([l^2_1,...,l^2_D])$ and depends on the characteristic length-scale $l_i$.

For making a prediction given input set $x_*$, the mean output value and variance could be represented as:
\begin{align}
        m_f(x_*) &= k^T(X,x_*)(K(X,X)+\sigma^2_nI)^{-1}y=k^T(X,x_*)\beta \nonumber\\
        V(x_*) &= k(x_*,x_*)- \nonumber\\
        &\,\,\,k^T(X,x_*)(K(X,X) +\sigma^2_nI)^{-1}k(X,x_*).
    \label{eq:19}
\end{align}

We used control input and simulated position as input to GPR, and the difference between our dynamic model and real experiment data as output, which can be described in the following equation:
\begin{align}
    \begin{split}
        x_i&=(L_1,L_2,L_3,L_4, p_{sim})\\
        f(x_i) &= p_{sim}-p_{exp} = e_i\\
    \end{split}
    \label{eq:20}
\end{align}

Where $p_{sim}$ could either generated from our learned dynamic neural network or directly from simulated mathematical model. 

Different from tendon force control, all the tendon length control inputs have to be previously calculated in the tendon-length control dynamic Cosserat rod model to ensure fitting the shape of the continuum robot. During this process, we also obtained the simulated position data at each point. 


Furthermore, manually correcting the input to RNN at each discontinuity is cumbersome, making the usage of simulated position data obtained from the mathematical model a preferred choice. Since our learned dynamic model is trained from a mathematical model, this approach is equivalent to modelling error between our trained dynamic model and real robot. The output of GPR will be the value of the predicted mean calculated in Equation \eqref{eq:19} when given a test input.

 We have collected 1000 real points across workspace in GPR training. When a small number of points are needed, we sparsely select random points with larger intervals between them while maintaining the overall shape of the workspace. To prove the robustness of the GPR model, for the case of 100 real points, we first divide 1000 by 100 to get ten points for each cluster of points, then we randomly select 1 point from 10 points in each cluster. The cross-validation error for a GPR model fitting 1000 real data points is around 0.002 meters.
 
\subsection{Inverse Mapping of the Continuum Robot}
We have trained a similar inverse mapping of continuum robot using simulation data presented in \cite{george2020first}. The training process is similar to forward dynamic model training with a single RNN layer with 30 neurons and a linear output layer using 10000 simulated data :
\begin{align}
    \begin{split}
        (x_i, x_{i+1})\rightarrow u_i\\
    \end{split}
    \label{eq:21}
\end{align}

where $x_i$ and $x_{i+1}$ are the current position and next target position, $u_i$ is the control input. After training of 698 epochs, the validation mean absolute error is around 0.0024 meters. All the data points are taken from the same dataset with modification of the data order to train this dynamic inverse mapping model. To predict given a target position $x_{target}$, we represent it as:
\begin{align}
    \begin{split}
        (x_i, x_{target})\rightarrow u_i\\
    \end{split}
    \label{eq:22}
\end{align}

\subsection{Control Policy A}

The inverse mapping of the continuum robot is trained with simulation data, while there is still a sim-to-real gap to fill. For the controller, we have added an error term $e_{gp}$ obtained from GPR to correct the behaviour of our trained inverse mapping:
\begin{align}
    \begin{split}
        e_{gp} = f((L, x_{s-target}))\\
        x_{s-target} = x_{real\: target} + e_{gp}\\
    \end{split}
    \label{eq:23}
\end{align}

Given a real target point $x_{real\: target}$ in the real workspace, now we need to estimate the target point $x_{s-target}$ in the simulated workspace. Term $e_{gp}$ is also dependent on the input value of  $x_{s-target}$. It then becomes an optimisation problem of finding suitable $x_{s-target}$ value. We take an initial guess of $x_{s-target} = x_{real\:  target}$. feeding it into the RNN and receive tendon length control outputs. Then feeding the control outputs and $x_{s-target}$ into the GPR model to generate $e_{gp}$. This error term will be added to $x_{s-target}$ as a correction of initial guess.

For open-loop control, the initial position does not change during the control time step. However, for closed-loop control, the initial position is constantly updated according to the time step. Meanwhile, we expect the current position in the simulated workspace instead of the real workspace as an input for our inverse mapping model. To simplify the optimisation step, we used the error predicted $e_{gp}$ from the GPR model at the simulated target point and treating it as an error term for each current position, adding that to the real position feedback input gives us a simulated current position. Thus, as the robot gets closer to the simulated target point, the error term becomes more accurate. The detailed algorithm is presented in Algorithm 2.

\begin{algorithm}[t]
\caption{Optimize $x_{s-target}$ with iterative approach in closed-loop control}
\label{alg:2}
\begin{algorithmic}
\Require $x_{real\: target}, x_{inital}$, time-step: $i$
\State initial guess: $x_{s-target} = x_{real\:target}$
\While{$\left \| x_{s-target}(i-1) - x_{s-target}(i) \right \| > 0.001 $}

\State Get control input  $(x_{initial},x_{s-target}(i-1))\rightarrow L(i)$
\State Predict using GPR $e_{gp}(i)=f((L(i),x_{s-target}(i-1)))$
\State Obtain tip position $x_{real(i)}$ from  tracking system
\State Update current position $x_{inital} = x_{real}(i) +  e_{gp}(i)$
\State $x_{s-target}(i) = x_{real\: target}+e_{gp}(i) $

\EndWhile
\end{algorithmic}
\end{algorithm}

\subsection{Control Policy B}

 Given the GPR model trained from real experiment data, we can generate $e_{gp}$ using Equation \eqref{eq:20} for each simulated input. Assuming each $e_{gp}$ is accurate, we will then obtain accurate real workspace positions by using the following equation to generate approximated real data:
\begin{align}
    \begin{split}
    p_{exp} = p_{sim}-e_{gp}
    \end{split}
    \label{eq:24}
\end{align}

With approximated real workspace points, we can retrain the RNN in control policy A using Equation \eqref{eq:21}. One benefit of using this method is to allow the RNN to learn the real dynamics of the continuum robot, similar to training the RNN with real experimental data. After training of 76 epochs, the validation means absolute error is around 0.0019 meters fitting 10000 generated data points.

\subsection{Control Policy C}

This control policy takes advantage of both control policy A and B, and combines them into a hybrid model. Control policy A relies on an accurate simulation model, and the GPR model compensates for the error. Most of the time, the simulation model is a good approximation of the real robot, and it works well when the simulation model is accurate.

Due to simplifications and many assumptions in the simulation model, some physics could not be captured well. This problem becomes more critical for our application, where soft materials are used. We use control policy B to learn the physics missing or heavily distorted in the real robot from the same GPR model generated data. While the approximated data generated from the GPR model could be less accurate than real data, it will outperform control policy A when the simulation model is not accurate. 

We have presented the control policy C to connect policy A and B using the final position tracking error as a criteria and checking $L_A$ and $L_B$ to avoid an infinite loop. If the absolute position error is greater than a constant value $e_c$, we will switch to the control output generated from policy B assuming the simulation model is not accurate at this location as presented in Figure \ref{fig:7}. And $e_c$ is chosen based on the average goal-reaching error of policy A : 0.015m. Currently, policy C relies on feedback to check $e_c$, works in the closed-loop setup.

\section{Experimental Results And Discussion}
We conducted experiments using the setup in Figure \ref{fig:3}. The control frequency is set to 5Hz and data collection of the tip position on the continuum robot from motion tracking systems output at a frequency of 5Hz. 
\begin{figure}[b!]
 \centering
 \includegraphics[width=8cm,height=4cm]{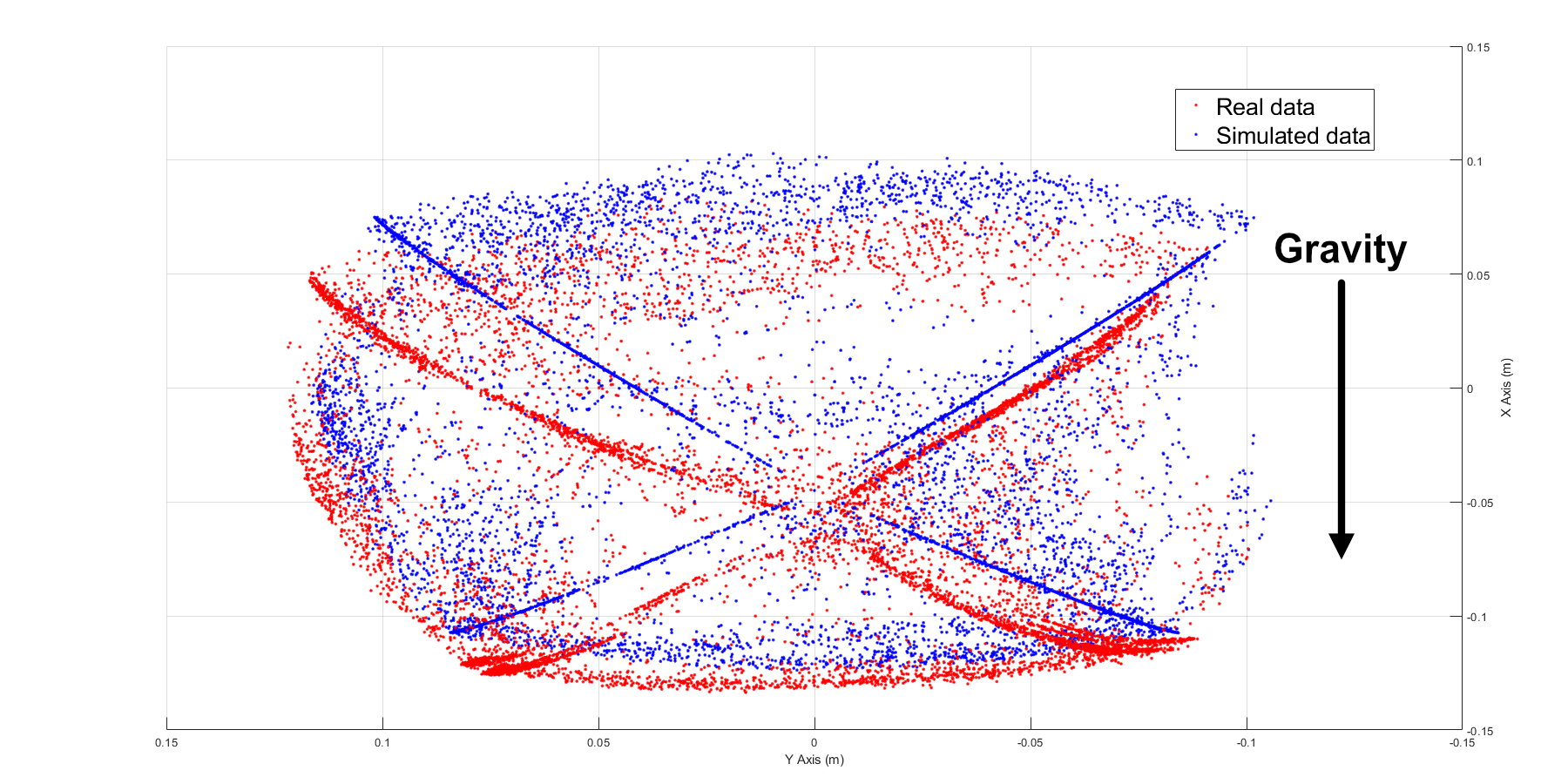} 
 \caption{Comparison of data generated from simulation and data collected from real experiments. Blue dots represent simulated data and red dots are real experiment data}
 \label{fig:5}
\end{figure}
\subsection{Comparison of the Simulated Workspace with the Real Workspace}

In the tendon-length control Cosserat dynamic model, we have generated 10000 data points. Using the same control inputs, we perform experiments with all these 10000 points on the real robot. We then present the difference between the simulated workspace and the real workspace in Figure \ref{fig:5}. 

After calibration, the error in z-axis is not apparent. However, due to the gravitational effect in the negative x direction and our continuum robot's unique triangular shape spine design, the error in the direction of the x-axis is significant where gravitational force is in the direction of negative x-axis.

\subsection{Determine Training Data Size for Gaussian Process}
\begin{figure}[b!]
 \centering
 \includegraphics[width=8cm,height=12cm]{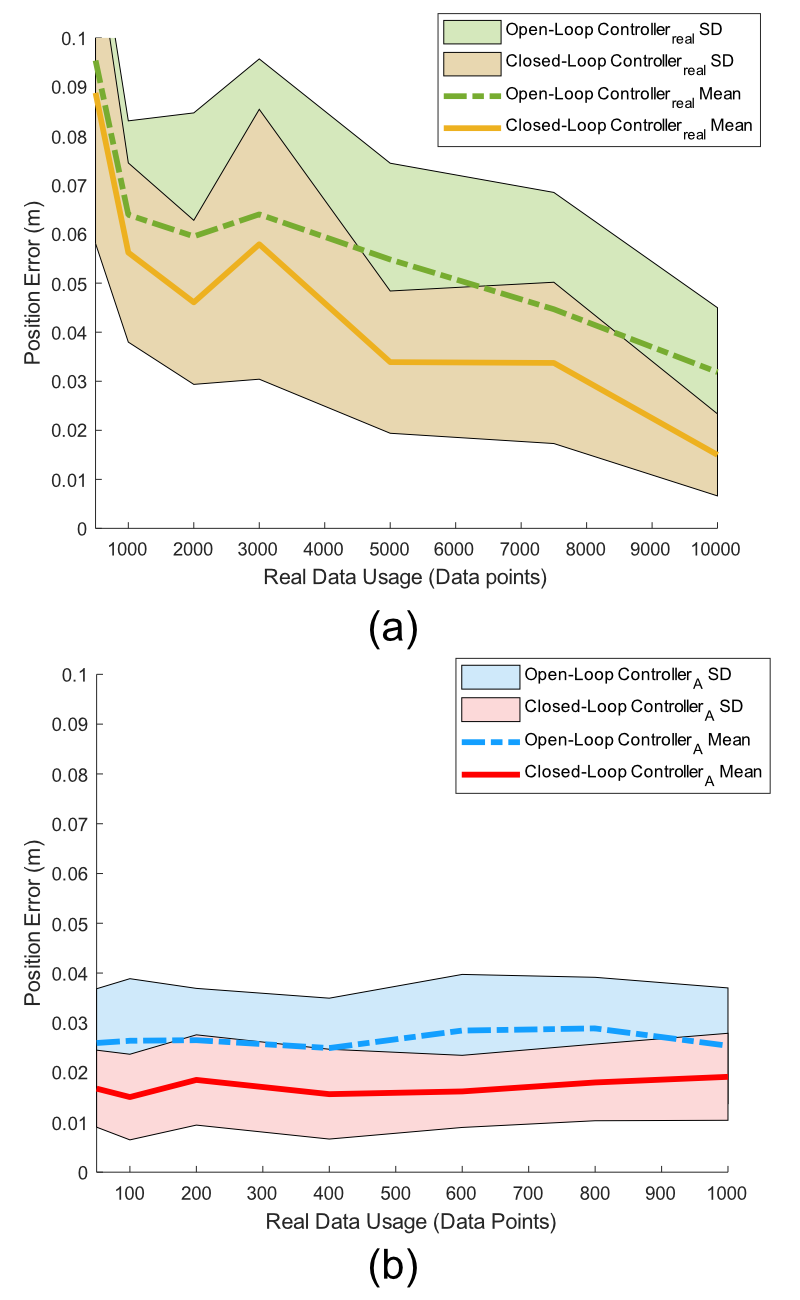} 
 \caption{(a) Comparison of Controllers trained with only real data using different number of real data points, (b)  Control policy A with the different GPR models trained with real data from 50 to 1000 points with the same RNN trained using 10000 simulation points, And their performance comparison with standard deviation plots respect to real data usage in goal reaching experiments}
 \label{fig:6}
\end{figure}
To test the capability of different controllers, we have experimented with our controllers by randomly picking 50 target points from another 960 unique points spaced equally in the real workspace and recording their position error. All the points selected for testing are not used for the training or validation process of the neural network or the GPR model. 

While we have collected 1000 pre-defined real data points for GPR training, we sparsely random-picking a certain number of them by maintaining the relative equal distance between each point. We have performed experiments with 50, 100, 200, 400, 600, 800, 1000 of real data to train GPR models and test our controller's  performance to find out the impact of real training data on our GPR models and controllers. Meanwhile, controllers that trained from only real data with 500, 1000, 2000, 3000, 5000, 7500 and 10000 real data points have been tested in the open/closed-loop case. The results are shown in Figure \ref{fig:6}.

For the controller trained directly with real data, more data points did increase the performance and decrease standard deviation error.

As for our model-based controller with GPR, fitting more real data in GPR does not necessarily improve the performance of our controller. The main reason could be the more uncertainties and errors added to the GPR model when fitting more data points. On the other hand, performing an inversion of matrix in GPR with a larger dataset costs $O(N^3)$. We find that a train data size around 100 is a good train data selection criteria for modelling the errors between dynamic model and real robot while maintain low standard deviation error.

\subsection{Goal Reaching Error}
\begin{table}[b]
\begin{threeparttable}
\caption{Performance Comparison for Goal reaching for different policies}
\label{tab:2}
\setlength\tabcolsep{0pt} 

\begin{tabular*}{\columnwidth}{@{\extracolsep{\fill}} ll cccc}
\toprule
     Controller & \multicolumn{3}{c}{Goal Reaching Error (m)} \\ 
\cmidrule{2-4}
     & Open-loop    &    Closed-loop   &    Closed-loop with load \\
\midrule
     A & 0.02640(0.01247) & 0.01510(0.00859) & 0.02962(0.01185)\\
\addlinespace
     B & 0.02875(0.00960) & 0.01546(0.00654) & 0.02878(0.00987)\\
\addlinespace
     C & N/A & 0.01091(0.00461) & 0.02807(0.00961)\\
\addlinespace
     $\pi_{real}$ & 0.03186(0.01314) & 0.014987(0.00837) & 0.03279(0.01277)\\

\bottomrule
\end{tabular*}

\smallskip
\scriptsize

\RaggedRight
$\pi_{real}$ Control Policy trained directly from real data.\\
\label{table:1}
\end{threeparttable}

\end{table}

The previous section found that using 100 real points trained GPR model is data-efficient while having good performance. This section presents control policy A, B, and C performance using the GPR model trained from only 100 real data points. Here, We compare our three control policies in Figure \ref{fig:7} with controller $\pi_{real}$ directly trained from 10000 real data using Equation \eqref{eq:21}. The performance with standard deviation is presented in Table \ref{table:1}. Control policy B has been retrained using the same GPR model with 10000 generated data points.

We can find that the closed-loop control improves the accuracy for control policy A, B, and controller $\pi_{real}$ compared to open-loop control. Both control policy A and B achieve comparable performances compared to controller $\pi_{real}$. Control policy A is slighter better than control policy B. Control policy C has outperformed controller $\pi_{real}$ with average goal-reaching an error of 1.1cm and a lower standard deviation error of 0.46cm. The uncertainties due to the soft spine deformation and long recovery time from stress aside from dynamic effects cause controller $\pi_{real}$ to be less accurate. With model-based learning with GPR framework, we reduced the uncertainties in data and improved the performance of our controller.

In addition, we have tested the same controllers performance when a tip-load of 20 grams is added to the soft continuum robot. In this case, the learned dynamic models from simulation and real data mismatched with real robot dynamics due to significant deformation of the continuum robot. Our proposed control policy A, B and C still outperformed the controller $\pi_{real}$.

\subsection{Closed-loop Path Tracking}

\begin{figure}[t]
 \centering
 \includegraphics[width=8cm,height=17cm]{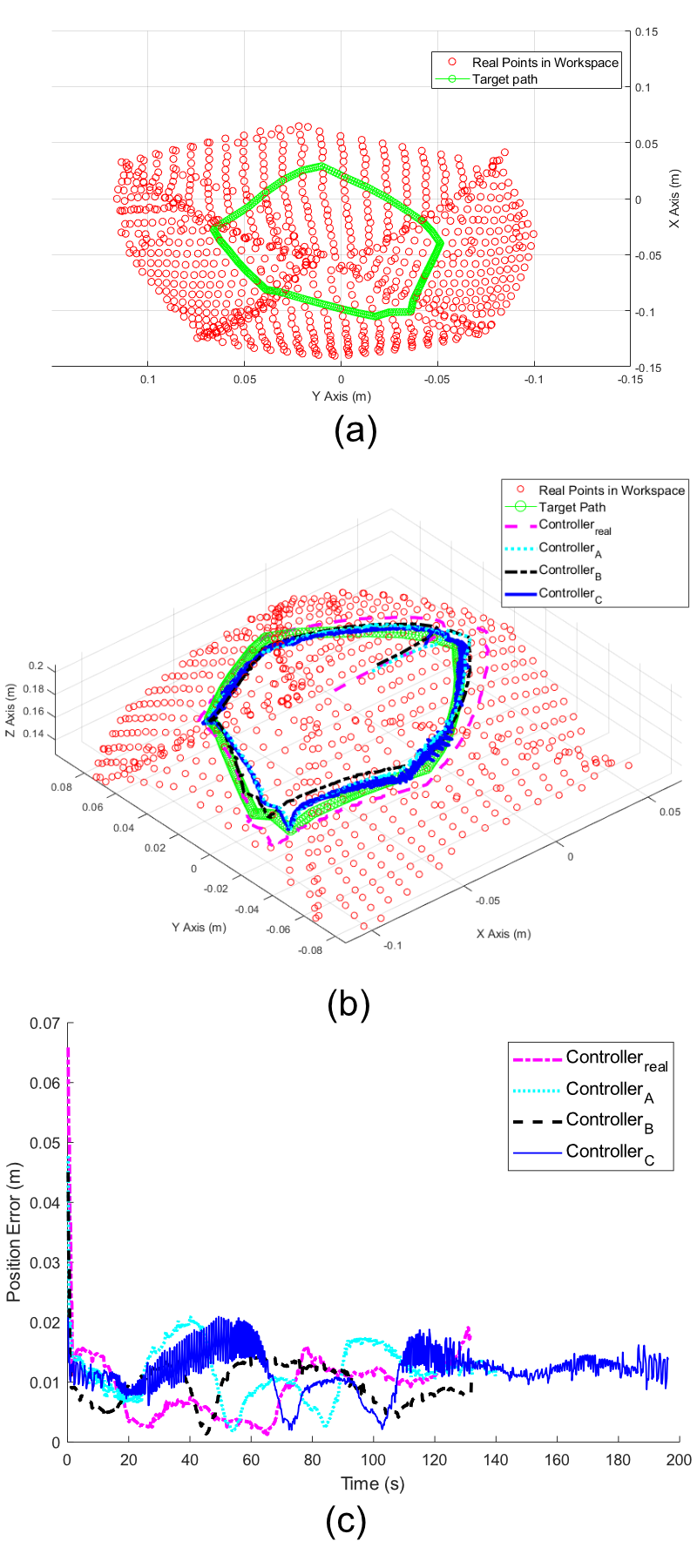} 
 \caption{(a) Target Path within real workspace (b) Path Tracking results for control policy A, B, C and $\pi_{real}$ (c) Time series plot of path tracking error.}
 \label{fig:8}
\end{figure}
We have connected 33 real data points in the workspace with linear interpolation to generate a target path as shown in Figure \ref{fig:8} (a) to perform closed-loop path tracking experiments. The resulted path tracking plots showed both control policy A and B are stable and closely match the target path with a completion time of around 135, 150 seconds, respectively. Furthermore, the overall path tracking error is similar for control policy A, B, C and $\pi_{real}$. The proposed control policy C is a hybrid model of policy A and B. It will switch between policy A or B based on a preset criteria, this approach results in a small oscillation of 3-5mm around the target path and longer completion time of 196 seconds. However, control policy C has much smaller starting point position error than other control policies, proving its advantage in goal reaching tests. 

\section{Conclusion}
We have proposed the data-efficient model-based learning framework with Gaussian process regression to perform tendon-length control on a continuum robot with compressible spines. Using only 100 pre-defined real data points, we could outperform the controller directly trained from real data in goal reaching tests with hybrid control policy C and comparable results with control policy A and B. Control policy A and B, on the other hand, is more stable than policy C in path tracking experiments.

For our model-based control framework, we could further improve their performance under external load by retraining RNN with simulation data knowing the exact amount of tip-load, while controller $\pi_{real}$ needs to recollect thousands of real data points. Control policy C's oscillation case during path tracking is due to frequent switching between control policy A and B. More complex criteria could be added to avoid frequent switching by analysing large error regions where the trained simulation model is inaccurate compared to real robot, and sticking to control policy B in that region. While our controller utilises a relatively simple inverse mapping of the continuum robots, the proposed framework is suitable for any model-based learning method and applicable to different types of continuum robots relying on cosserat rod model. The obtained dynamic model could perform either optimisation for planning or model-free policy learning using reinforcement learning to obtain complex control policies. Future work will focus on these aspects, specifically modelling external load effects.





\section*{Acknowledgment}
The authors would like to thank Angus B. Clark for his help in the development of the experimental set up. 

\bibliography{bibliography}
\bibliographystyle{unsrt}

\end{document}